\documentclass[9pt]{article}
\usepackage{spconf,amsmath,graphicx}
\usepackage{IEEEtrantools}

\usepackage{enumitem}
\usepackage{xcolor}
\usepackage{url}

\usepackage[font=small,labelfont=bf]{caption}
\setlist{nosep, leftmargin=14pt}

\begin{document}

\bstctlcite{BSTcontrol} 

\begin{titlepage}

    \huge
    \noindent
    {\bfseries IEEE Copyright Notice\par}
    \vspace{1.5cm}

    {\large
    ©2026 IEEE. Personal use of this material is permitted. Permission from IEEE must be obtained for all other uses, in any current or future media, including reprinting/republishing this material for advertising or promotional purposes, creating new collective works, for resale or redistribution to servers or lists, or reuse of any copyrighted component of this work in other works.
    \par}

    \vspace{1cm}

    \large
    \noindent
    {\itshape
    Accepted for publication, 2026 IEEE 23rd International Symposium on Biomedical Imaging (ISBI), April 2026, London, United Kingdom. DOI: to appear.
    \par}

\end{titlepage}

\title{\Large Disease Progression and Subtype Modeling for Combined Discrete and Continuous Input Data}

\name{%
\shortstack{%
Sterre de Jonge$^{1}$,
Elisabeth J.~Vinke$^{1,2}$,
Meike W.~Vernooij$^{1,2}$,
Daniel C.~Alexander$^{3}$,\\
\textit{Alexandra L.~Young$^{3*}$ and Esther E.~Bron$^{1*}$}%
}\thanks{$^{*}$Authors contributed equally.}
}

\address{%
\normalsize $^{1}$ Department of Radiology and Nuclear Medicine, Erasmus MC, Rotterdam, The Netherlands\\
\normalsize $^{2}$ Department of Epidemiology, Erasmus MC, Rotterdam, The Netherlands\\
\normalsize $^{3}$ Hawkes Institute, Department of Computer Science, University College London, London, United Kingdom
}

\maketitle

\ninept

\begin{abstract} 
Disease progression modeling provides a robust framework to identify long-term disease trajectories from short-term biomarker data. It is a valuable tool to gain a deeper understanding of diseases with a long disease trajectory, such as Alzheimer's disease. A key limitation of most disease progression models is that they are specific to a single data type (e.g., continuous data), thereby limiting their applicability to heterogeneous, real-world datasets. To address this limitation, we propose the Mixed Events model, a novel disease progression model that handles both discrete and continuous data types. This model is implemented within the Subtype and Stage Inference (SuStaIn) framework, resulting in Mixed-SuStaIn, enabling subtype and progression modeling. We demonstrate the effectiveness of Mixed-SuStaIn through simulation experiments and real-world data from the Alzheimer's Disease Neuroimaging Initiative, showing that it performs well on mixed datasets. The code is available at: \url{https://github.com/ucl-pond/pySuStaIn}. 
\end{abstract}


\section{Introduction}
\label{sec:intro}
Disease progression modeling provides a robust framework to identify long-term disease trajectories from short-term biomarker data \cite{young_data-driven_2024,yoshioka_disease_2024}. This is particularly valuable for neurodegenerative disorders, such as Alzheimer's disease (AD), where progression remains poorly understood, but the disease trajectory has a long preclinical course. Insights into disease trajectories are crucial for the development of targeted interventions for specific patient subpopulations. Various progression models have been applied to neurodegenerative disorders to gain a deeper understanding of underlying biological heterogeneity \cite{young_uncovering_2018,vogel_four_2021,salvado_disease_2024,venkatraghavan_disease_2019,venkatraghavan2025,koval_ad_2021}.


One of the earliest disease progression models is the event-based model (EBM), which conceptualizes disease progression as a series of events. In EBM, biomarker values transition from a normal to an abnormal value, and the model estimates the most likely ordering of these transitions (i.e., events) \cite{fonteijn_event-based_2012,young_data-driven_2014,venkatraghavan_disease_2019}. Biomarkers are expected to follow a bimodal distribution, distinguishing disease from reference populations. Other disease progression models have been developed to accommodate ordinal input data \cite{young_ordinal_2021,poulet_multivariate_2023} or continuous input data \cite{young_uncovering_2018}. While early models focused on a single disease trajectory, recent approaches estimate multiple trajectories to account for phenotypic heterogeneity, in addition to temporal heterogeneity \cite{young_uncovering_2018,venkatraghavan2025,koval_ad_2021}. 

A key limitation of these disease progression models is that they model a fixed trajectory shape. This restricts the types of data to which they can be applied and limits their applicability across diverse datasets. Table \ref{tab:sustain-models} summarizes the existing models that can be embedded in Subtype and Stage Inference (SuStaIn) \cite{young_uncovering_2018}, illustrating the different suitable input data types corresponding to the underlying biomarker trajectory. Integration of multiple data types, however, could offer a more complete view of disease trajectories. Within the context of AD, this may provide additional insights across changes in structural imaging, clinical scores, or visual ratings.

We therefore propose the Mixed Events model, a novel disease progression model that handles discrete and continuous input data types. The key innovation is a novel formulation of the likelihood function that enables integration of the likelihoods of the piecewise linear \textit{z}-score model, the event-based model, and the scored events model into a single objective function for estimating the event ordering. We integrate this model into SuStaIn, leading to Mixed-SuStaIn, enabling subtype and progression modeling for mixed datasets. We validate our approach through experiments using synthetic data and data from the Alzheimer's Disease Neuroimaging Initiative (ADNI), demonstrating its effectiveness in obtaining characteristic orderings across diverse biomarker data types.

\vspace{-2mm}

\begin{table}[h]
\centering
\scriptsize
\caption{Existing disease progression models embedded in SuStaIn.}
\label{tab:sustain-models}

\setlength{\tabcolsep}{3pt} 
\renewcommand{\arraystretch}{1.3} 

\begin{tabular}{|p{1.4cm}|p{3.2cm}|p{3.2cm}|}
\hline
\textbf{Model} & \textbf{Biomarker Trajectory} & \textbf{Input Data Type} \\ \hline

Event-based model &
Biomarkers transition from a normal to an abnormal value reflecting an instantaneous switch in disease progression &
Biomarker values transformed to probabilities based on a bimodal data distribution (normal vs. abnormal) \\ \hline

Scored events model &
Biomarkers transition from a normal score (e.g., value zero) to higher scores, reflecting an instantaneous switch between levels of disease severity &
Ordinal biomarker scores transformed to probabilities, where each biomarker value in each individual is assigned a probability for each possible score \\ \hline

Piecewise linear \textit{z}-score model &
Biomarkers linearly increase from one \textit{z}-score to another &
Biomarker values transformed to \textit{z}-scores based on a Gaussian reference distribution \\ \hline

\end{tabular}
\end{table}

\vspace{-3mm}

\section{Methods}
\subsection{Mathematical Model for Mixed Data}
\begin{figure}[h]

\begin{minipage}[b]{1.0\linewidth}
  \centering
  \centerline{\includegraphics[width=6cm]{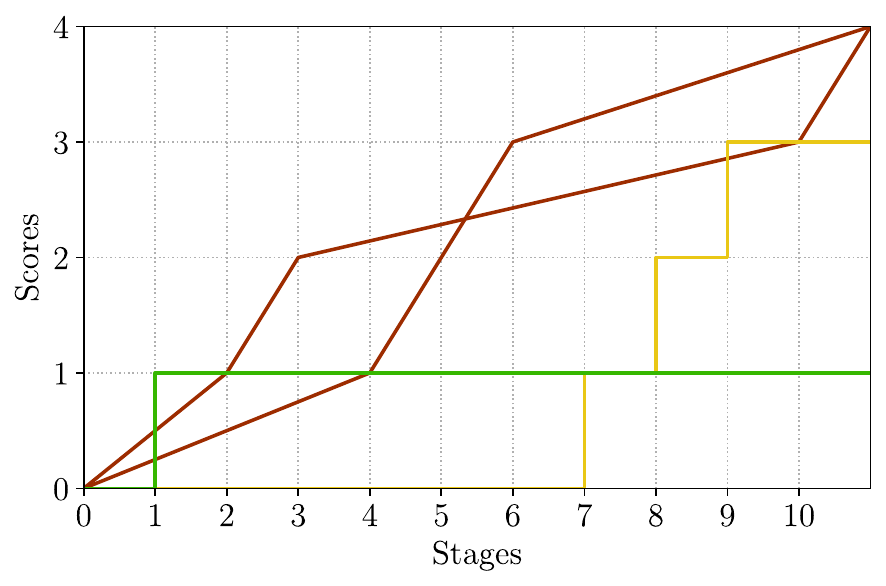}}
\end{minipage}
\caption{Example biomarker trajectory. \textit{Z}-scored biomarkers (red) reach abnormality at \textit{z}-scores 1–3 and accumulate at \textit{z}-max = 4. The ordinal biomarker (yellow) reaches abnormality at scores 1–3, while the binary biomarker (green) transitions once, resulting in ten events.}
\label{fig:biomarker-trajectories}

\end{figure}

The Mixed Events model is a generalized disease progression model for mixed data that describes disease progression as a sequence of binary (normal to abnormal), ordinal and \textit{z}-score events. At each model stage, a biomarker transitions from one level to another according to the chosen model type for that biomarker. To illustrate the different biomarker trajectories given a sequence, we visualized example trajectories in Figure \ref{fig:biomarker-trajectories}, where \textit{z}-scored biomarkers (red) follow a piecewise linear trajectory, and binary (green) and ordinal (yellow) biomarkers follow a step-wise trajectory. 

\noindent The likelihood of the data given a sequence of biomarker events, where each event represents a particular biomarker reaching a new level or score, can be described by

\begin{equation} \label{eq:generalized-likelihood-sequence}
    P (X_j |S) = \sum_{k=0}^{K} P (k) \prod_{i=1}^{I} P (x_{i,j} |S, k, M(i))
\end{equation}

\noindent where $X_j$ represents the vector of data points $x_{i,j}$ for biomarkers $i=1 ... I$ for each subject $j=1 ... J$, $S$ is the sequence of biomarker events, $k$ is the stage along the sequence $S$, and $M(i)$ is the chosen model for biomarker $i$. The model $M(i)$ can be binary ($B$), ordinal ($O$), or \textit{z}-score ($Z$). The code is available at: 
\url{https://github.com/ucl-pond/pySuStaIn}.

\vspace{2mm}
\noindent
\textbf{Binary (event-based) model likelihood} The binary (event-based) model and the ordinal (scored events) model share the assumption that a biomarker transitions in discrete steps to reflect disease progression. In the binary model \cite{fonteijn_event-based_2012}, each biomarker shifts from a normal to an abnormal value at a specific disease stage. If the event has not yet occurred at a given stage $k$, the likelihood is given by $p(x|\neg E_i)$, whereas if the event has occurred, the likelihood is given by $p(x|E_i)$. The probabilities for the normal distribution and the abnormal distribution are typically estimated using Gaussian mixture modeling \cite{fonteijn_event-based_2012,young_data-driven_2014} or kernel density estimation \cite{firth_sequences_2020}, although any probability distribution can be used. The likelihood for the binary model ($B$), can be written as 

\begin{equation}
    P(x_{i,j}|S,k,M(i)=B) = \begin{cases} P(x_{i,j}|E_{i}) & \text{ for } s(i) \le k \\ P(x_{i,j}|\neg E_{i}) & \text{ for } s(i)>k \end{cases} 
\end{equation}

\noindent where $s(i)$ describes the stage at which biomarker $i$ becomes abnormal in the sequence $S$.

\vspace{2mm}
\noindent
\textbf{Ordinal (scored-events) model likelihood} In the ordinal (scored events) model \cite{young_ordinal_2021}, ordinal biomarkers transition to a progressively higher score. The probability that score $w$ is reached by biomarker $i$ at stage $k$ can be modeled by any probability distribution defined by the user, e.g., a categorical distribution \cite{young_ordinal_2021} or a Gaussian distribution. The likelihood for the ordinal model ($O$) is formulated as

\begin{equation}
    P(x_{i,j}|S,k,M(i)=O) = P(x_{i,j}|E_{s(i,w)})
\end{equation}

\noindent where $s(i,w)$ describes the biomarker $i$ and score $w$ that are reached at stage $k$. The binary model can also be described in this form by setting the scores $w$ to $w = (\text{normal},\text{abnormal})$, and so the ordinal model is a generalization of the binary model.

\vspace{2mm}
\noindent
\textbf{Z-score (piecewise linear) model likelihood} In the z-score (piecewise linear) model \cite{young_uncovering_2018}, the ordering of events is described as the linear accumulation of biomarkers increasing from one z-score to another. The likelihood for z-scored ($Z$) biomarkers, is given by

\begin{equation}
     P(x_{i,j}|S,k,M(i)=Z) = \text{NormPDF}(x_{i,j},g_{s(i,w)}(k),\sigma_{i})
\end{equation}

\noindent where the input data $x_{ij}$ are provided as \textit{z}-scores which need to be transformed to probabilities to compute the likelihood for a given sequence. The data value is compared to the point estimate of the piecewise linear trajectory for \textit{z}-score \textit{z} at stage $k$ \cite{aksman_pysustain_2021}. Originally, it was compared to the integral over the piecewise linear trajectory, but it was found that the point estimate, which is more efficient, showed similar results \cite{young_uncovering_2018,aksman_pysustain_2021}. The \textit{z}-scores accumulate at \textit{z}-max at stage $k+1$, both $z$-values and \textit{z}-max are defined by the user.

\subsection{Subtyping}
The previous sections described fitting a single disease progression trajectory. In SuStaIn, multiple subtypes $C$ are fitted with distinct trajectories. The likelihood for Mixed-SuStaIn can be written as

\begin{equation}
    P(X|S) = \prod_{j=1}^{J}\sum_{c=1}^{C}P(c)P(X_j|S_c)
\end{equation}

\noindent with $P(X_j|S_c)$ the Mixed Event likelihood function from Eq. \ref{eq:generalized-likelihood-sequence}. We determined the optimal number of subtypes $C$ with five-fold cross-validation \cite{young_uncovering_2018}. 

\subsection{Simulation Experiments}
We evaluated the stability of the proposed method under different simulation settings. Synthetic data for each input type were generated using previously described methods (\textit{z}-scored: \cite{young_uncovering_2018}, ordinal: \cite{young_ordinal_2021} and binary: \cite{fonteijn_event-based_2012}). Different configurations were tested, evaluating the number of subjects (250, 500$^\star$, 1000), number of subtypes (1, 3$^\star$, 5), number of biomarkers (2+1+1, 4+2+2$^\star$, 6+3+3) (\textit{z}-scored + ordinal + binary biomarkers) and two settings for biomarkers values (events) (1: {[}1,2,3{]} + {[}1,2,3{]} + {[}1{]}$^\star$, 2: {[}1,3,5{]} + {[}1,3,5{]} + {[}1{]}) (\textit{z}-score + ordinal + binary values). Default parameters ($^\star$) were used for the other variables. Each experiment was repeated ten times for different randomly chosen subtype progression patterns and simulated datasets. Performance was assessed using Kendall's rank correlation. 

\subsection{Real-World Data Validation}
For validation on real-world data, we used the Alzheimer's Disease Neuroimaging Initiative (ADNI) database\footnote{\url{http://adni.loni.usc.edu}, \url{http://www.adni-info.org}}. ADNI is an open-access database containing imaging, clinical, and biomarker data from subjects diagnosed with AD, mild cognitive impairment (MCI), and cognitively normal (CN) individuals (reference group).

\vspace{2mm}
\noindent
\textbf{Data} The ADNIMERGE dataset was downloaded on Feb 3, 2025. We included all participants with baseline 3T MRI and cerebrospinal fluid (CSF) data, resulting in 641 subjects (209 CN, 341 MCI, 91 AD). Clinical diagnosis 24 months after baseline served as an outcome measure. Total brain, ventricles, hippocampus, entorhinal cortex, middle temporal gyrus, and fusiform gyrus (extracted with FreeSurfer v5.1) were included as continuous biomarkers in the Mixed Events model. Volumes were \textit{z}-scored using CN participants as the reference group and corrected for age and intracranial volume, and log-transformed for ventricles. The \textit{z}-values and \textit{z}-max were derived from the data by defining the \textit{z}-values as the set of integers up to the 95\% quantile and \textit{z}-max as the 99\% quantile rounded to the nearest integer. The CSF measures amyloid-$\beta_{1-42}$, phosphorylated tau (p-tau) and total tau (t-tau) were included as binary biomarkers in the model, as these measures exhibit relatively well-defined thresholds that separate normal from abnormal values. The CSF measures were log-transformed, followed by estimation of the probabilities belonging to the normal ($p(x|\neg E_i)$) and abnormal ($p(x|E_i)$) distribution with Gaussian mixture modeling \cite{young_data-driven_2014}.

\vspace{2mm}
\noindent
\textbf{Comparison Analysis} We benchmarked the performance of Mixed-SuStaIn against EBM-SuStaIn, which is the only previously-developed disease progression model that can include both discrete and continuous biomarkers. EBM-SuStaIn achieves this by limiting the modeled trajectory shapes to normal and abnormal transitions and treating the input data as binary (B) events. For both methods, the same number of subtypes was used. 

To compare methods, we first evaluated predictive performance for subjects converting from CN to MCI and from MCI to AD, using AUC-ROC. Baseline patient staging (Figure \ref{fig:patient-staging}; also depicting converters by stage) was used to predict conversion within 24 months. Additionally, we evaluated the relationship between SuStaIn stage and cognition by correlating stage with Mini-Mental Stage Examination (MMSE), stratified by subtype, using Pearson's correlation.

\section{Results}
\subsection{Simulation Experiments}
\begin{figure}[h]
    \centering
    \includegraphics[width=0.85\linewidth]{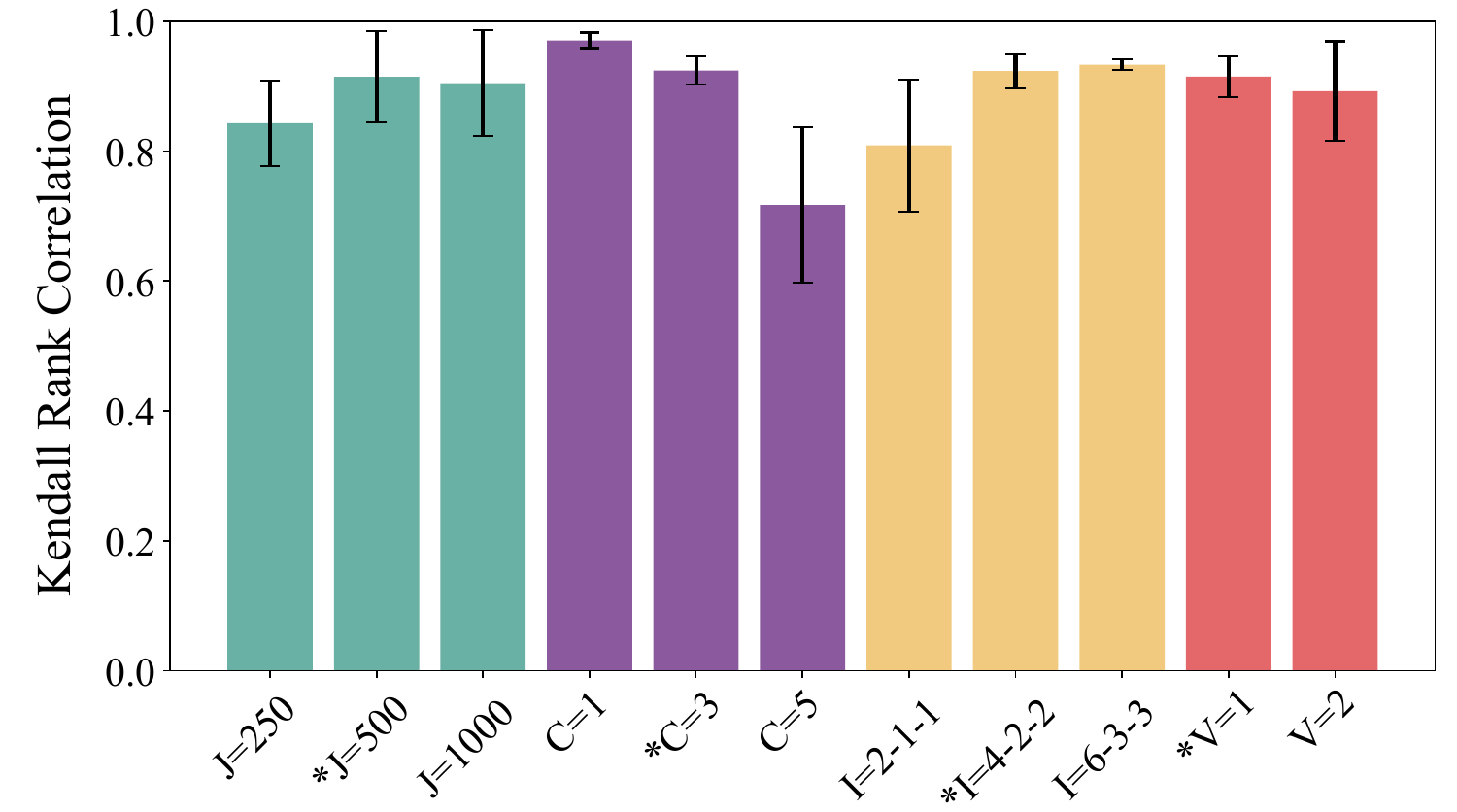}
    \caption{Accuracy of Mixed-SuStaIn in recovering the ground truth subtype patterns on synthetic data. Error bars indicate standard deviation. Experiments settings: number of subjects ($J$, green), subtypes ($C$, purple), biomarkers ($I$, yellow) and values ($V$, red). $^\star$ indicates default values.}
    \label{fig:sim-results}
\end{figure}

Simulation experiment results are shown in Figure \ref{fig:sim-results}. All settings achieved a Kendall rank correlation in the range of 0.6-1, indicating that Mixed-SuStaIn was able to recover the ground truth subtype patterns well. Increasing number of subtypes $C$, reduced the Kendall rank correlation, as there are fewer subjects per subtype. Decreasing number of biomarkers $I$ also reduced the Kendall rank correlation, which may be explained by the fact that subtype trajectories are more strongly defined and separated by larger number of biomarkers. 

\subsection{Real-World Data Validation}
\begin{figure*}[ht]
    \centering
    \includegraphics[width=0.9\textwidth]{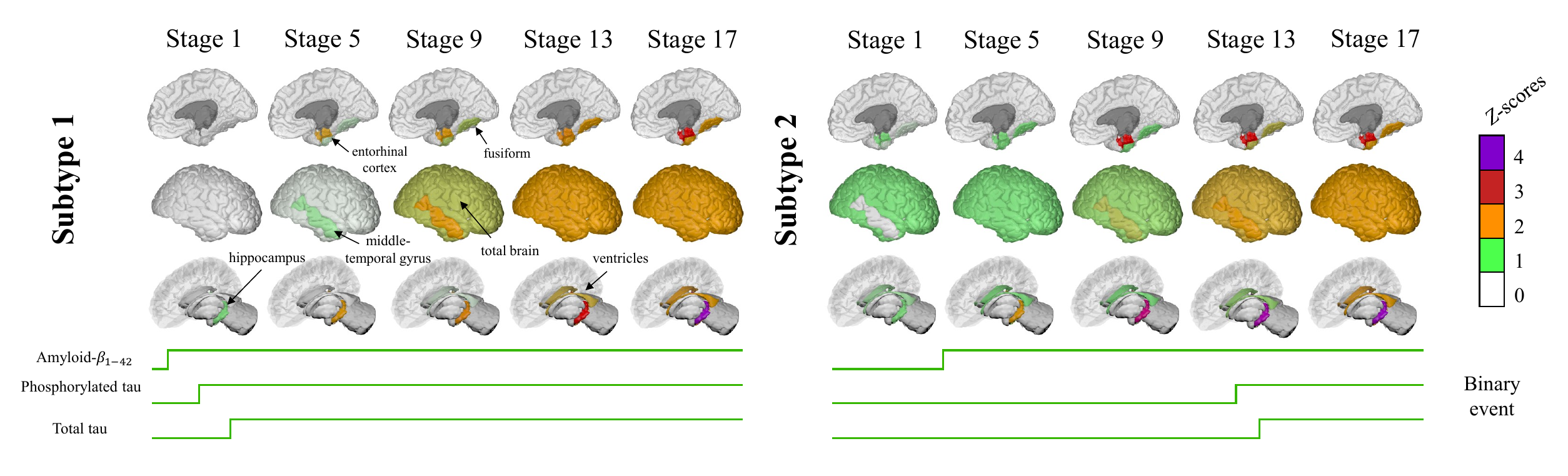}
    \caption{Disease progression patterns of subtype 1 (n=458) and subtype 2 (n=183) identified by Mixed-SuStaIn. The top rows show cortical regions becoming increasingly abnormal (higher \textit{z}-scores) across disease stages. The “total brain” biomarker reflects global brain change, but is visualized only on cortical areas for clarity. Bottom rows depict binary progression of cerebrospinal fluid biomarkers.}
    \label{fig:pos-var_mixed-sustain}
\end{figure*}

\begin{figure}[h]
    \centering
    \includegraphics[width=0.93\linewidth]{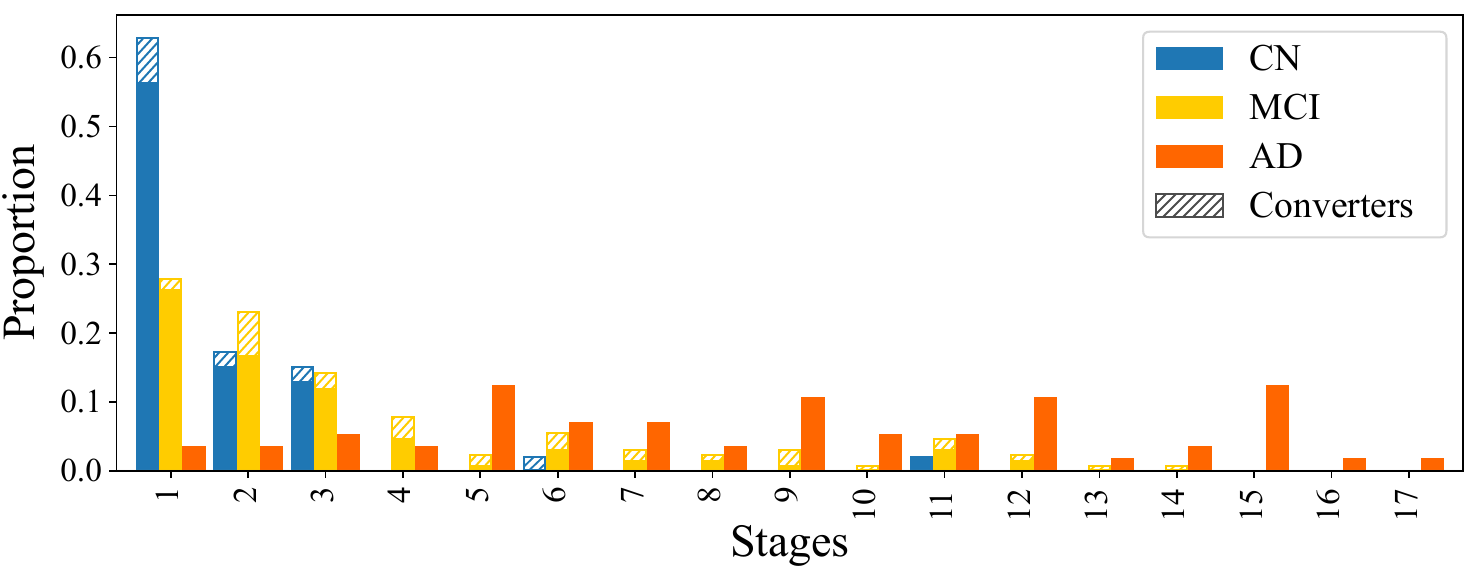}
    \centerline{(a) subtype 1}\medskip

    \includegraphics[width=0.9\linewidth]{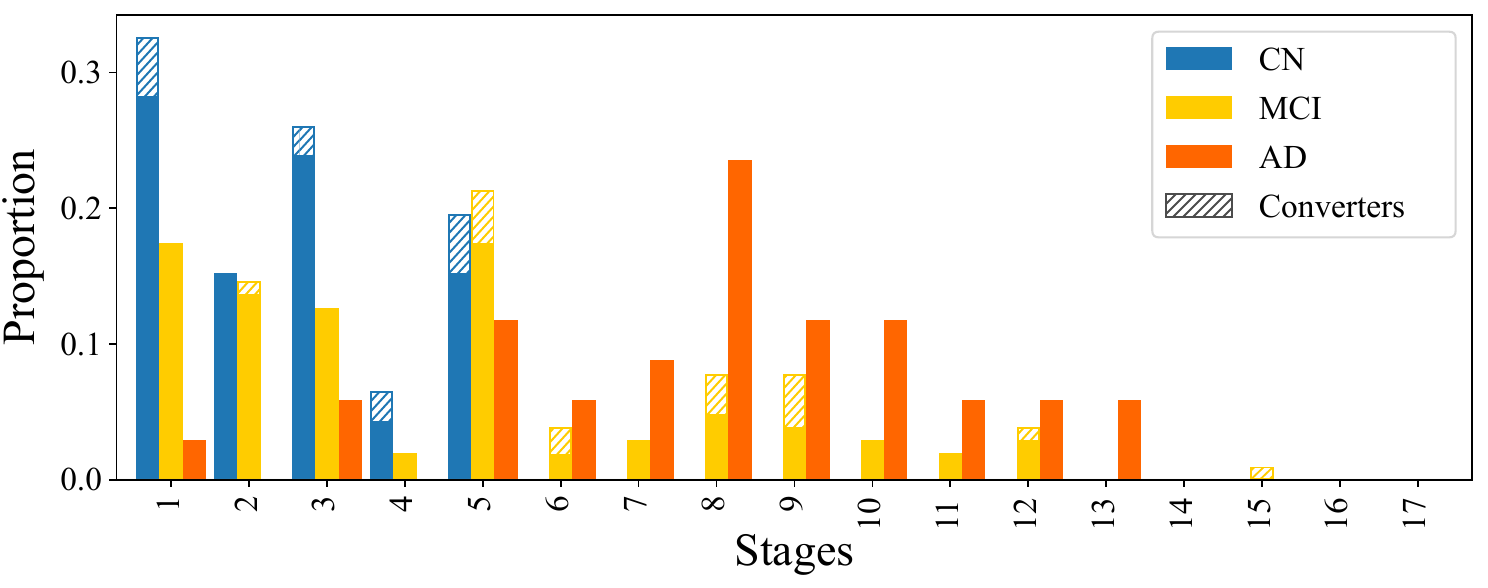}
    \centerline{(b) subtype 2}\medskip

    \caption{The probability that subjects from each diagnostic category, including the proportion of converters, belong to each Mixed-SuStaIn stage for subtype 1 (a) and subtype 2 (b). Converters in the CN bars represent CN-to-MCI conversion, converters in the MCI bars represent MCI-to-AD conversion. AD=Alzheimer's Disease, CN=cognitively normal, MCI=mild cognitive impairment.}
    \vspace{5mm}
    \label{fig:patient-staging}
\end{figure}

Study subject demographics and input biomarker characteristics are summarized in Table \ref{tab:demographics}. Brain regions were modeled with the following \textit{z}-score ranges (\textit{z}-max in parentheses): hippocampus, 1–4 (5); entorhinal cortex, 1–3 (5); total brain, middle temporal gyrus and fusiform, 1–2 (4); and ventricles, 1-2 (3). The optimal number of subtypes was identified as two. The disease progression patterns for the subtypes are shown in Figure \ref{fig:pos-var_mixed-sustain}. Subtype 1 (n=352) followed a typical AD progression, with amyloid-$\beta_{1\text{--}42}$, p-tau, and t-tau becoming abnormal first. Subtype 2 (n=289) was characterized by early atrophy in the hippocampus, total brain, and entorhinal cortex, followed by amyloid-$\beta_{1\text{-}42}$ and late tau pathology. This subtype appears to capture subjects who deviate from the typical AD progression pattern seen in subtype 1. A similar 'cortical subtype' was reported by Estarellas et al. \cite{estarellas2024}. The subject staging for both subtypes shows that CN and MCI subjects were generally assigned lower disease stages than AD patients (Figure \ref{fig:patient-staging}). Converters (within 24 months) were proportionally assigned higher stages, indicating a correspondence between disease stage and clinical progression. 


\vspace{-2mm}

\begin{table}[ht]
\caption{Baseline demographics, likelihood of belonging to the abnormal distribution, and regional brain \textit{z}-socres for included subjects grouped by diagnostic category: cognitively normal (CN), mild cognitive impairment (MCI), and Alzheimer's disease (AD).}
\centering
\footnotesize
\setlength{\tabcolsep}{5pt} 
\renewcommand{\arraystretch}{1.1}
\begin{tabular}{|l|c|c|c|}
\hline
\textbf{Characteristic} & \textbf{CN (n=209)} & \textbf{MCI (n=341)} & \textbf{AD (n=91)} \\
\hline
\textit{Demographics} & & & \\
Female sex (\%) & 54.2 & 44.7 & 42.1 \\
Age (years), $\mu$ ± SD & 72.8 ± 5.8 & 71.2 ± 7.2 & 73.8 ± 8.5 \\
\textit{Likelihood, $\mu$ ± SD} & & & \\
Amyloid-$\beta_{1\text{--}42}$ &  0.28 ± 0.35 & 0.47 ± 0.38 & 0.81 ± 0.26 \\
Total tau & 0.12 ± 0.19 & 0.19 ± 0.27 &  0.43 ± 0.31 \\
Phosphorylated tau &  0.25 ± 0.27 & 0.35 ± 0.34 & 0.68 ± 0.30 \\
\textit{Z-scores, $\mu$ ± SD} & & & \\
Total Brain & 0.0 ± 1.0 & 0.32 ± 1.11 & 1.34 ± 1.15 \\
Ventricles & 0.0 ± 1.0 & 0.30 ± 1.11 & 0.83 ± 1.10 \\
Hippocampus  & 0.0 ± 1.0 & 0.94 ± 1.45 & 2.40 ± 1.11 \\
Middle temporal gyrus & 0.0 ± 1.0 & 0.30 ± 1.10 & 1.60 ± 1.17 \\
Entorhinal cortex & 0.0 ± 1.0 & 0.58 ± 1.33 & 1.98 ± 1.19 \\
Fusiform & 0.0 ± 1.0 & 0.27 ± 1.09 & 1.27 ± 0.97 \\
\hline
\end{tabular}
\label{tab:demographics}
\end{table}

\vspace{-2mm}

\noindent
The disease progression patterns identified by EBM-SuStaIn are described below. For the two subtypes, subtype one (n=458) showed typical AD progression, with the following ordering of events: amyloid-$\beta_{1-42}$, p-tau and t-tau, hippocampus, entorhinal cortex, middle temporal gyrus, fusiform, total brain and ventricles. The second subtype (n=183) followed the ordering: hippocampus, entorhinal cortex, amyloid-$\beta_{1-42}$, ventricles / total brain (same position), middle temporal gyrus, fusiform and lastly, p-tau and total-tau. 

In CN subjects, 158 remained stable within 24 months and 16 converted to MCI. In MCI subjects at baseline, 209 remained stable and 52 converted to AD. In prediction of conversion from CN to MCI, Mixed-SuStaIn achieved an AUC of 0.724 compared to 0.723 for EBM-SuStaIn. For conversion from MCI to AD, Mixed-SuStaIn achieved an AUC of 0.828 compared to 0.825 for EBM-SuStaIn. Overall, both methods had a similar performance in the prediction tasks.

For subtype 1, the correlation between SuStaIn stage and cognition was $r$ = -0.69 for Mixed-SuStaIn and $r$ = -0.63 for EBM-SuStaIn. For subtype 2, the correlations were $r$ = -0.43 (Mixed-SuStaIn) and $r$ = -0.41 (EBM-SuStaIn). In both models, the correlation was stronger in subtype 1, which reflects the dominant AD atrophy pattern, and weaker in subtype 2. Overall, the correlations were similar.

\section{Conclusion}
In this paper, we introduced the Mixed Events model, a novel disease progression model that integrates \textit{z}-score, ordinal and binary models into a single objective for estimating event orderings in mixed datasets. Embedded within SuStaIn, the model successfully recovered meaningful event orderings across mixed data types in real-world ADNI data. Mixed-SuStaIn demonstrated equal predictive performance and also showed similar correlations to cognition to the benchmark method. The key added value of the proposed approach, however, is the increased modeling flexibility for mixed datasets, for which no current alternative exists. In the comparison analysis, we only included biomarkers compatible with both models to enable a direct comparison, necessarily excluding ordinal biomarkers. This may have limited performance gains. In future work, we will apply Mixed-SuStaIn to heterogeneous, population-based data, including a broader range of biomarkers, to further evaluate its generalizability and demonstrate its ability to uncover progression patterns beyond the scope of single data-type models. 


\section{References}

\begingroup
  \renewcommand{\baselinestretch}{0.9}\footnotesize\selectfont
  \setlength{\parskip}{0pt}

  \makeatletter
  \renewenvironment{thebibliography}[1]
    {%
     \list{\@biblabel{\@arabic\c@enumiv}}%
          {%
           \usecounter{enumiv}%
           \settowidth\labelwidth{\@biblabel{#1}}%
           \leftmargin\labelwidth
           \advance\leftmargin\labelsep
           \setlength{\itemsep}{0pt}%
           \setlength{\parsep}{0pt}%
          }%
     \sloppy\clubpenalty4000\widowpenalty4000%
     \sfcode`\.=1000\relax
    }
    {\endlist}
  \makeatother

  \bibliographystyle{IEEEtran}
  \bibliography{references}
\endgroup

\end{document}